\documentclass[10pt,twocolumn,letterpaper]{article}

\usepackage{iccv}
\usepackage{times}
\usepackage{epsfig}
\usepackage{graphicx}
\usepackage{amsmath}
\usepackage{amssymb}
\usepackage{booktabs}
\usepackage{multirow}
\usepackage{array} 
\usepackage{subcaption}
\usepackage{float}
\usepackage[x11names]{xcolor}

\usepackage[pagebackref=true,breaklinks=true,letterpaper=true,colorlinks,bookmarks=false]{hyperref}

\iccvfinalcopy 

\def\iccvPaperID{4599} 

\ificcvfinal\fi
\newcommand*{\ShowNotes}{}
\ifdefined\ShowNotes
  \newcommand{\colornote}[3]{{\color{#1}\bf{#2: #3}\normalfont}}
\else
  \newcommand{\colornote}[3]{}
\fi

\definecolor{teaser_pink}{RGB}{190, 0, 0} 
\definecolor{teaser_green}{RGB}{101, 156, 64} 
\definecolor{teaser_blue}{RGB}{56, 102, 182} 
\definecolor{teaser_brown}{RGB}{159, 72, 15} 
\definecolor{teaser_gray}{RGB}{99, 99, 99} 
\definecolor{teaser_yellow}{RGB}{230, 173, 0} 
\definecolor{arrow_green}{RGB}{84, 130, 53}
\definecolor{arrow_red}{RGB}{192, 0, 0}
\definecolor{clavicula}{RGB}{127, 96, 0}

\begin{document}

\title{LIMITR: Leveraging Local Information for Medical Image-Text Representation}
\author{Gefen Dawidowicz \hspace{0.15in} Elad Hirsch \hspace{0.15in}  Ayellet Tal\\
Technion – Israel Institute of Technology\\}

\maketitle
\ificcvfinal\thispagestyle{empty}\fi

\begin{abstract}
Medical imaging analysis plays a critical role in the diagnosis and treatment of various medical conditions.
This paper focuses on chest X-ray images and their corresponding  radiological reports.
It presents a new model that learns a joint X-ray image \& report representation.
The model is based on a novel alignment scheme between the visual data and the text, which takes into account both local and global information.
Furthermore, the model integrates domain-specific information of two types---lateral images and the consistent visual structure of chest images.
Our representation is shown to benefit three types of retrieval tasks:
text-image retrieval, class-based retrieval, and phrase-grounding.
\end{abstract}

\section{Introduction}

 \begin{figure}[tp]
\centering
\includegraphics[width=\linewidth]{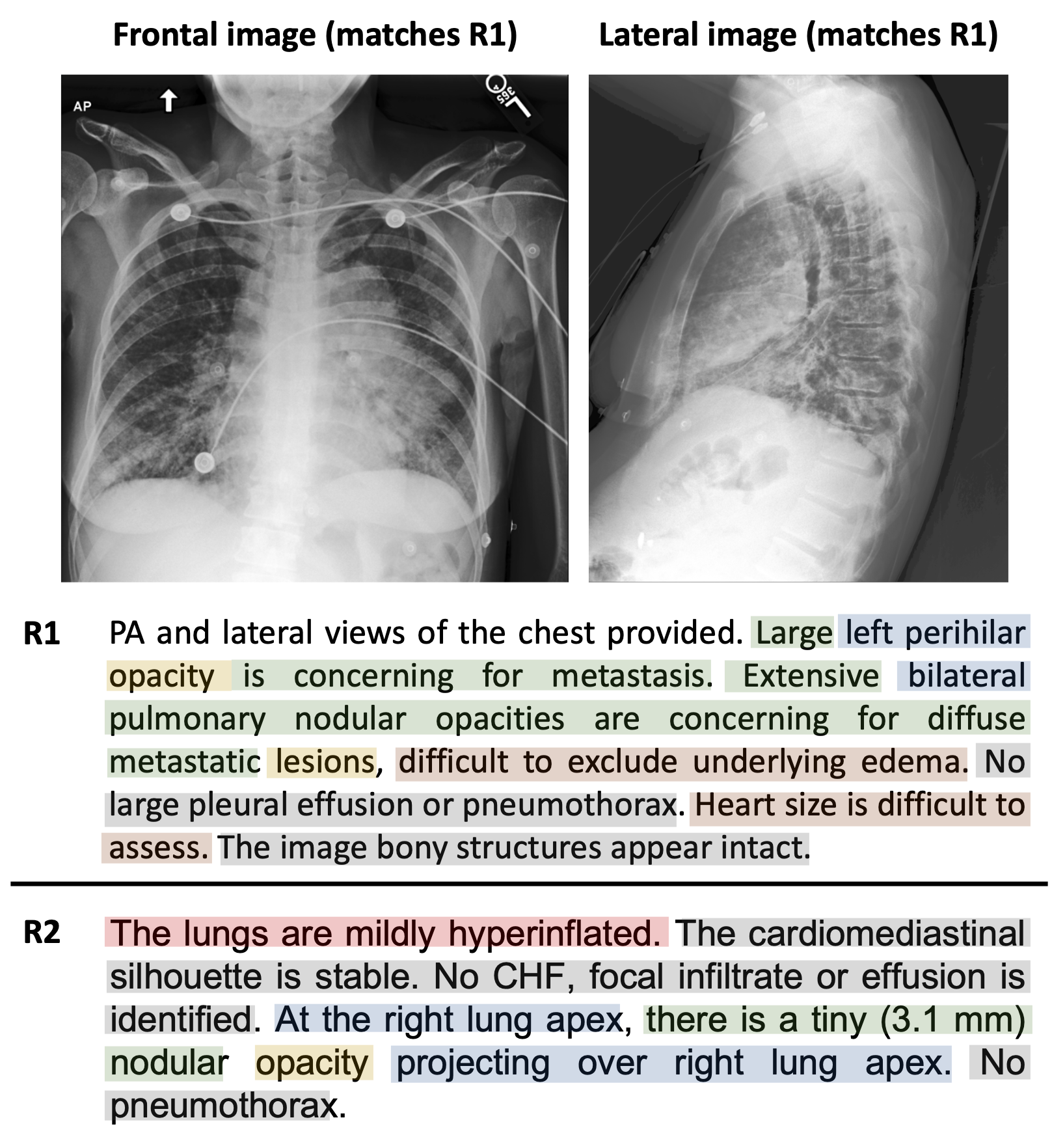}
\caption{{\bf Image-report retrieval.} 
Given the visual data, which contains  both Lung Opacity \& Lung Lesion diagnoses, our model retrieves the correct {\bf R1} report.
Another report, {\bf R2}, which corresponds to another image, contains the same \textcolor{teaser_yellow}{pathologies}.
It differs in the subtle details:
\textcolor{teaser_green}{pathology description}, \textcolor{teaser_blue}{localization}, \textcolor{teaser_brown} {uncertainties}, \textcolor{teaser_gray}{normalities} and additional \textcolor{teaser_pink}{unlabelled pathologies}.
In tasks such as  class-based retrieval, {\bf R2} is considered a correct match.
Our model aims also at tasks, such as image-text retrieval, which care about the subtleties.}
\label{fig:teaser}
\end{figure}

X-ray imaging is one of the most common medical examinations performed,  where in the U.S. alone $70$ million scans are performed every year~\cite{Smith-Bindman2008}. 
During a post-scan routine, a medical report is written by a radiologist. 
This is a challenging task  even for trained personnel, since the pathologies, which typically occupy a small portion of the image, might be characterized by subtle (sometimes distributed) anatomical changes. 
Automating the analysis has the potential to aid experts and to speed-up the process.

The task of producing a joint representation of medical image-report data inherently differs from that of natural image-text, which has been extensively explored recently~\cite{du2022survey}.
The available data is orders of magnitude smaller than that of natural images.
Furthermore, the data is highly unbalanced, since most of the examples are normal.
Even in abnormal examples, most of the image (/report) is normal.

Due to the importance of the task, 
the available medical data has been used for a variety of applications, including class-based retrieval (retrieving data of the same diagnosis)~\cite{huang,Zhang2020,Moon,wang2022medclip},
 pathology classification~\cite{wang2022multi,huang,Boecking_2022,Zhang2020,Moon,wang2022medclip},
detection~\cite{huang,Boecking_2022,wang2022multi,lovt},
and phrase-grounding~\cite{Boecking_2022}.
Often, these tasks do not require subtle details, such as the severity of the pathology, its exact location, or findings that are beyond the pre-defined pathologies.

We propose a novel model, which  learns a joint X-ray image-report representation that is attentive to subtle details, additional descriptions of the pathologies, uncertainties etc., as illustrated in Figure~\ref{fig:teaser}.
It is based on three key ideas. 
First, 
our method learns to utilize both global alignment (image-report pairs) that captures high-level information, such as the sheer existence of a disease,
and local alignments (region-word pairs) that capture subtle details and abnormalities.
Second, it benefits from lateral images, which are usually ignored.
Third, it utilizes domain-specific localization information. 
We elaborate below.

Local alignment in medical imaging is challenging since the data is not annotated locally by bounding boxes and their labels.
This is in contrast to natural image datasets, which provide localized ground-truth information~\cite{sgraf,
DBLP:journals/corr/abs-1712-02036,
DBLP:journals/corr/KarpathyF14,
DBLP:journals/corr/abs-1803-08024,
DBLP:journals/corr/abs-1909-02701,
DBLP:journals/corr/abs-2004-06165}.
Furthermore, localization ambiguity is inherent in medical imaging, as report findings may  correspond to multiple image regions. 
To this end, we propose a new aggregation method and a new loss, which synthesize both region-word alignments within a single pair and  global and local alignments across pairs.

Second, we propose to use lateral views, if they exist, just like radiologists do. 
These views may provide additional information, yet they are largely ignored by other representation learning works.
We introduce an attention model that learns for each portion of the report when to consider both images, when only one, and when none.

Lastly, our model utilizes some basic domain-specific localization information---the structure of the human body; for instance, the heart is always inbetween the lungs and is approximately in the same image position.
We show that we can  add global positional encoding for the whole dataset.
This encoding also allows the network to better learn local connections between the frontal and the lateral views.

To demonstrate the benefit of our model,  we evaluate it for three different retrieval tasks: 
(1)~{\em Text-image retrieval:} Given a report, the goal is to find the most suitable image and vice versa.
In this task, we expect the corresponding  report (/image) to be retrieved, as illustrated in Figure~\ref{fig:teaser}, where all the information that appears in the text and in the image, are taken into account.
This  task demonstrates the ability of our method to accurately capture subtleties.
(2)~{\em Phrase-grounding:} Given an image and a corresponding phrase, the goal is to produce an attention map of the similarity between the phrase and each region.
This task demonstrates the quality of the local alignment. 
(3)~{\em Class-based retrieval:} Given a textual description, the goal is to retrieve images that belong to the same class of the description.
This is the most common retrieval task.
We present SoTA results for all these tasks.

Hence, our work makes the following contributions:
\begin{enumerate}
\vspace{-0.05in}
    \item 
    We propose a novel model for learning a joint X-ray image \& report representation. 
    It is based on a new local and global alignment scheme.
    This alignment is shown to boost performance.
\vspace{-0.05in}
    \item
    Our model integrates additional domain-specific knowledge---lateral images and visual structure. 
    This information further improves performance. 
\vspace{-0.05in}
    \item
    The benefit of our model is demonstrated on three retrieval tasks, showing its ability to capture fine features in both images and reports.
    We demonstrate SoTA results on commonly-used datasets.

\end{enumerate}

\section{Related work}
\label{sec:related}

Recent works on text-image multi-modal representations for {\em chest X-ray (CXR)} datasets are used for a variety of tasks.
These tasks are either uni-modal or multi-modal, 
as briefly reviewed hereafter.

\vspace{0.05in}
\noindent
{\bf Uni-modal tasks.}
The common uni-modal tasks are visual, where the focus is on classification and localization.
Specifically, given an image, the goal of~\cite{Boecking_2022,huang,wang2022multi,wang2022medclip,Zhang2020} is to classify it into pre-defined diagnoses.
This can be done for multi-label classes~\cite{huang,wang2022multi,wang2022medclip,Zhang2020} or for binary classification~\cite{Boecking_2022,huang,wang2022multi,Zhang2020}.

In localization, we are given images and aim  to learn localized information for detection and segmentation tasks \cite{Boecking_2022,huang,lovt,wang2022multi}.
These tasks are designed for single-pathology studies and rely on relatively small datasets.
Examples include  RSNA pneumonia detection~\cite{RSNA}, foreign objects detection~\cite{objectCXR} and SIIM pneumothorax segmentation~\cite{SIIM}.

Textual analysis in this domain is less prevalent.
In~\cite{Boecking_2022}, the focus is on CXR domain-specific language understanding tasks, such as text classification and masked-token prediction.
The joint training with the images provides a superior language model for these tasks.

\vspace{0.05in}
\noindent
{\bf Multi-modal tasks.}
Our focus is on  multi-modal tasks, for which there exist less works.
In~\cite{huang,Zhang2020,wang2022medclip} the task is {\em class-based retrieval}, i.e. given an example in one modality (image/text), retrieve examples from the other modality. The requirement is that the retrieved examples should belong to the same class of diagnoses as the query.

In {\em phrase-grounding}, introduced by~\cite{Boecking_2022}, we are given an image and a corresponding phrase describing a pathology in the image, and aim to localize the image regions that match the phrase.
This is a challenging task as specific relations between phrases and  certain image features need to be learned.
However, this work ignores studies with multiple pathologies, as well as more elaborated phrases that express uncertainties and descriptions of normalities.

We introduce an additional  retrieval task, {\em text-image (or image-text) retrieval}, where the accuracy of the representation can be better evaluated.
We expect to retrieve the exact match from one modality, given the other.


 \begin{figure*}[tb]
\centering
\includegraphics[width=1\textwidth]{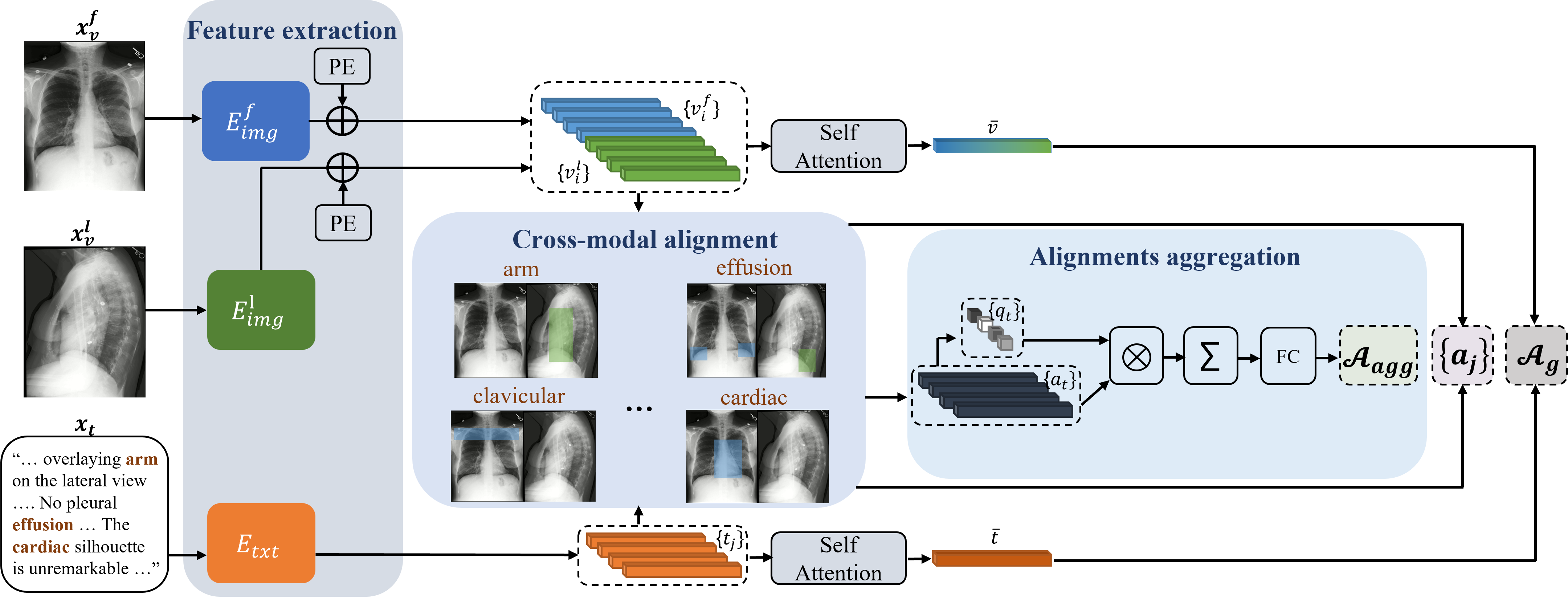}
\caption{{\bf Model.}
The model consists of three blocks.
(1)~For feature extraction, 
two CNNs,  $E_{img}^{l}$ \& $E_{img}^{f}$ for the lateral~\&  the frontal images, are used as the image encoders and one pre-trained network, $E_{txt}$, is used as the text encoder.
The local visual representations are concatenated and a global representation is created using self-attention.
A global representation is similarly created for the report.
The distance between these representations forms our global alignment $\mathcal{A}_g$, which is used in our global loss, $L_g$.
(2)~During cross-modal alignment, alignments between the local representations of the two modalities are calculated (Figure~\ref{fig:cross modal alignment}). 
These alignments, $\{a_j\}$,
are used for our local internal loss $L_{int}$ (Figure~\ref{fig:local internal loss}).  
Thanks to our attention mechanism, the model works well with or without lateral images.
(3)~The local alignments are aggregated using learned significance scores, to create the final image-report similarity score, $\mathcal{A}_{agg}$. 
These scores are 
used in our local external loss~$L_{ext}$ (Figure~\ref{fig:local external loss}).
} 
\label{fig:model}
\end{figure*}

\section{Model}
Our goal is to learn an informative joint representation of X-ray images and their reports.
In this joint representation space, an image and its corresponding report should be mapped to close points, whereas mismatched pairs should reside farther apart.
Recall that our model realizes three key ideas.
First, to account for the different importance of the local image regions (/report words) to the global alignment,  we propose a novel aggregation method of the local representations, which incorporates learned  importance.
Second, our model is the first to use lateral images for representation learning, in addition to the frontal ones, when available.
Third, our model leverages the fact that medical images are characterized by a unique and known structure of the human body.

Our model, which is outlined in Figure~\ref{fig:model}, consists of three parts: 
(1)~feature extraction, which produces the local and the global representations, (2)~cross-modal alignment, which utilizes these features, as well as additional information, in order to find the alignments between the two modalities,
and (3)~local-alignment aggregation, which produces global alignment of an image and a report.
Our model optimizes the following loss (which will be elaborated upon in Section~\ref{subsec:loss}):
\begin{equation}
L=L_g+L_{ext}+L_{int}.
\label{eq:loss}
\end{equation}
Here,
the global alignment is optimized by
$L_g$, 
the local alignment of regions \& words across examples is optimized by
$L_{ext}$, and the local alignment of regions \& words within a single example is optimized by $L_{int}$.

\subsection{Feature extraction}
\label{subsec:feature}
Given an image  $x_v$ and its corresponding report $x_t$,
the visual and the textual features, are independently extracted.
Recall that in medical imaging, we lack local annotated data (bounding boxes and their labels); furthermore, the images suffer from  localization ambiguity, as findings 
may correspond to multiple image regions. 
Thus, pre-trained object detectors are not as useful as in the natural domain.

\noindent
\textbf{Visual feature extraction.}
We benefit from the localized nature of the intermediate layers of CNNs ($E_{img}$ in Fig.~\ref{fig:model}), to obtain $N_r$ region-level visual features $\left \{ v_1,...,v_{N_r} \right \}$. 
In particular, we use the output of the last {\em convolution} layer.

Next, we enrich the extracted features with our knowledge of the layout of the human body.
Specifically, in chest X-ray the organs are approximately in the same positions (which guide  radiologists in the diagnosis process).
We leverage this structure by encoding and integrating it within learning. 
We realize it through adding positional encoding.
Since our positional encoding is inherent in the input, it conceptually differs from that being used in {\em transformers}, which encode the relative positions.
We sum each visual local feature vector $v_i$ with a corresponding vector that encodes its spatial $2D$ position.
In our implementation we use the $2D$ sinusoidal encoding of~\cite{vit}, as follows.
Let $v_i$ be the $i^{th}$-patch features in the image and  $(x,y)_i$ be the  coordinates of this patch. 
Our visual feature vector is then defined as the sum of the visual encoder output with the positional encoding corresponding to the patch location in the image
$v_i\leftarrow v_i+PE((x,y)_i)$,
where $PE((x,y)_i)$ is  that of~\cite{vit}.

We observe that not all local regions are as important.
Specifically, some small regions---those that contain abnormalities---should count more than others.
To this end, we apply the self-attention operator of~\cite{vaswani2017attention} on the local features, in order to extract our global image feature representation, $\bar{v}$.
This $\bar{v}$  captures the relationships between all the local representations of a given image $x_v$.

In our implementation we use {\em Resnet-50}
~\cite{resnet} as the image encoder $E_{img}$, as it has been previously shown to benefit  medical images~\cite{Boecking_2022,huang,Zhang2020}. 

\noindent
\textbf{Frontal and lateral information.}
Lateral images, which exist in $50\%$ of the studies, contain information that may improve diagnosis, yet they are largely ignored.
Our approach utilizes the information from the lateral images, by learning to weigh information from both views.
In addition, as studies do not necessarily contain the lateral view, our model learns from studies with a single view, as well as from studies with both views.
Specifically, for each view type, we train a separate image encoder, $E_{img}^{l}$ and $E_{img}^{f}$. 
The features from the two encoders are concatenated to form the set $\left \{ v_1^f,...,v_{N_r}^f,v_1^l,...,v_{N_r}^l \right \}$ and are inserted to the cross-modal alignment module. 
When a lateral image is not available, we set $\left \{v_i^l \right \}_{i=1} ^{N_r} = \underline{\mathbf{0}}$. 

\noindent
\textbf{Textual features extraction.} 
Given a report $x_t$, it is first tokenized,  using the  vocabulary 
of~\cite{DBLP:journals/corr/abs-1904-03323}.
The sequence of tokens is then fed into a text encoder, $E_{txt}$.
This yields local textual features, at the token level, $\left \{ t_1,...,t_{N_w} \right \}$.
Lastly, since in analogy to image regions, certain words are more important than others, we apply self-attention to extract the global textual features, $\bar{t}$.

In our implementation, we use BioClinicalBERT~\cite{DBLP:journals/corr/abs-1904-03323} as our text encoder, due to its high performance.

\subsection{Cross-modal alignment}
\label{subsec:cross modal alignment}

\begin{figure}[tb]
\centering
\includegraphics[width=0.98\linewidth]{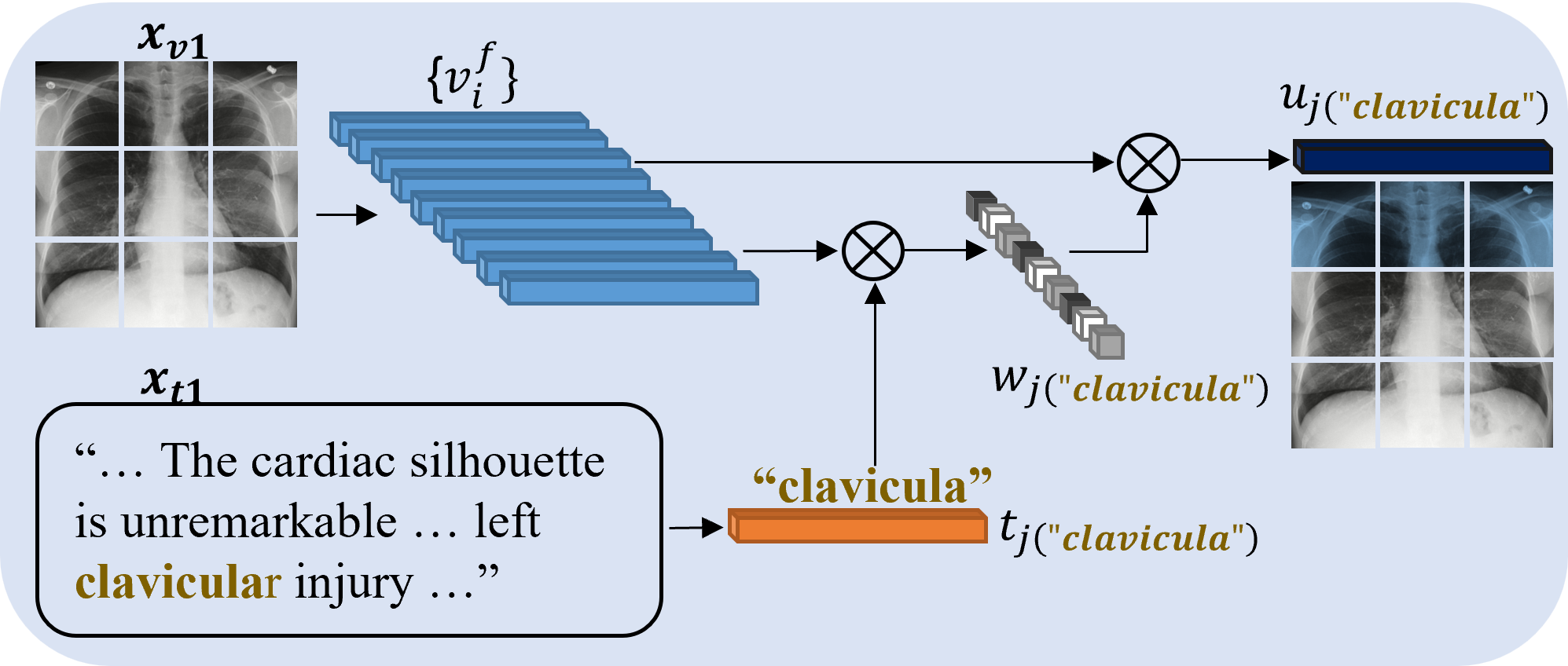}
\caption{{\bf Cross-modal alignment.} 
Given an image-report pair $(x_v,x_t)$, we compute for each word $t_j$ (e.g. \textcolor{clavicula}{clavicula}) its corresponding visual weighted representation $u_j$.
This is done by using the  similarity between each region representation $v_i^f$ to $t_j$ as the weight for this region $w_j[i]$. 
The right image shows that the regions that correspond to  "clavicula" are highlighted in blue, representing the higher weight of these regions.
The final visual representation, $u_j$, is created by a weighted sum of $v_i^f$.
}
\label{fig:cross modal alignment}
\end{figure}

\label{subsec:alignment}

Given a visual and textual local  representations,  
 $\left \{ v_1,...,v_{N_r} \right \}$ and $\left \{ t_1,...,t_{N_w} \right \}$ respectively,
our goal is to learn the alignments between the two modalities, including 
the alignments between report words and image regions.
To do that, our model should address two challenges:
(1)~the lack of local annotations connecting  the two modalities, 
(2)~the pathologies may occupy only a small area of the image, as well as a small part of the report. 
To this end, a fine-grained approach is sought-after.

Our approach, which is illustrated in Figure~\ref{fig:cross modal alignment}, takes into account the global representations of images and reports, as well as the local representations of regions and words.
This is done both by weighing the image region features with respect to each textual feature and vice versa---by weighing the textual features with respect to each region feature.
For clarity, in the following we will describe only how this is done in the first case (weighing image regions with respect to text).
However, in the loss, both are summed.

First, the cosine similarity $c_{ij}$ between $v_{i}$ and $t_{j}$ is computed, to create $c_j=[c_{1j},c_{2j},\dots,  c_{N_rj}]$.
It is further
normalized using softmax,  in order to get an attention weight:
\begin{equation}
{w _{j}=softmax(\lambda c_{j})}.
\end{equation} 
The attended visual feature $u_j$ with respect to the $j^{th}$ word is the weighted sum of all the visual local representations: 
\begin{equation}
{u_{j}=\sum_{i=1}^{N_r}w _{j}[i] \cdot v_{i}}.
\label{eq:uj}
\end{equation} 

Next, we calculate the alignment between $t_j$ and its corresponding  $u_j$. 
The local alignment $a_{j}$ is calculated as:
\begin{equation}
{a_{j}=\mathcal{A}(u_{j},t_{j})=\frac{ u_{j}\circ t_{j}}{\left \|  u_{j}\circ t_{j} \right \|_{2}}},
\label{eq:aj}
\end{equation}
where $\circ$ is an element-wise multiplication and $\left \| \cdot  \right \|_{2}$ is the $L_2$-norm.
Note that unlike~\cite{huang,lovt,wang2020covid}, our alignment $a_{j}$  is a vector rather than a scalar (that averages the entries). 
This is important for our aggregation method, as will be discussed in Section~\ref{subsec:aggregation}, where we learn the importance of the different $a_{j}$'s.
Our alignments $\left \{a_{j}\right \}$ will be later used for the local alignment loss $L_{int}$.

Similarly to the local alignments, we compute the global alignment, given the global image feature vector, $\bar{v}$ and the  global report feature vector, $\bar{t}$. 
It is computed as:
\begin{equation}
{\mathcal{A}_{g}=\mathcal{A}(\bar{v},\bar{t})=\frac{ \bar{v}\circ \bar{t}}{\left \|  \bar{v}\circ \bar{t} \right \|_{2}}}.
\label{eq:Ag}
\end{equation}
This $\mathcal{A}_{g}$ will be later used for the global loss $L_g$.

The cross-modal alignment is computed between each  region in both the frontal and the lateral images to each token of the report.
Hence, if significant information appears in the two images, regions from both receive high weights.
But, when important information appears only in one of the images, only its regions will get high weights. 
In a sense, our model mimics the radiologists actions.
When a pathology is available in the two views, they examine both;
otherwise, they focus on the relevant view.

\subsection{Aggregation}
\label{subsec:aggregation}
Our goal is to aggregate all the local alignments of a given image-report pair, in order to create the final representation of the pair, $a_f$.
We do not wish to simply sum the alignments~\cite{huang}, 
since not all regions should be treated equally.
In particular, most image regions and their corresponding report descriptions are normal observations, shared by all examples in the dataset, whereas the pathologies mostly  occupy small regions. 
The distinct information, e.g. the pathology, should be given more weight.

To account for this variance, we suggest to weigh the local alignments in accordance with their informativeness.
Let the local alignments, computed at the alignment module, be $\mathcal{A}_{T}=\left \{a_{1},a_{2},a_{3}...a_{N_w}\right \}$ (Equation~\ref{eq:aj}).
These alignments are aggregated into a single alignment vector using a weighted sum.
The weight of each vector is computed by self-attention, where (through the interaction between the local regions) it is determined which should  be more attended, in order to better perform the task.
Thus, for each~$a_t$ we compute its weight relative to the set $\mathcal{A}_T$.
Let $\bar{a}$ be the mean of $\mathcal{A}_T$.
The weight of $a_t$  is defined as:
\begin{equation}
q_{t}=\biggl( softmax \Bigl( \frac{W_q  \, \bar{a} \cdot (W_k \, \mathcal{A}_T)^T}{\sqrt{d}} \Bigr) \biggr)_t,
\end{equation} 
where $1 \leq t \leq N_w$, $W_q$ and $W_k$ are linear transformations of the self-attention, and $d$ is the feature dimension. 

The final alignment vector between an image and a report is defined as:
\begin{equation}
{a_{f}=\sum_{t=1}^{N_w} q _{t} \, (W_v \cdot a_{t})}.
\end{equation}
\label{sec:model}
This aggregated alignment vector represents the image-report pair.
It is  passed through an FC layer to produce the final scalar alignment score $\mathcal{A}_{agg}$, which will be later used for the external loss  $L_{ext}$.

 \subsection{Loss}
 \label{subsec:loss}

Recall that our goal is to maximize the similarity between positive image-report pairs and minimize the similarity between negative pairs.
To achieve this, we use three instances of the contrastive loss function~\cite{DBLP:journals/corr/abs-1807-03748}, each expressing a different concept: one global and two local, as seen in  Equation~\ref{eq:loss}.
The latter two enable different regions or words to be considered as having different significance.
We elaborate on these losses hereafter.

The global loss, $L_g$, attempts to maximize the global alignment $\mathcal{A}_g$ (Equation~\ref{eq:Ag}) of positive image-report  pairs and minimize the global alignment of negative pairs.
Let $(x_v^{k},x_t^{k})$ be a corresponding pair and $\tau$ be a temperature parameter. 
The loss is defined as:

\begin{equation}
\begin{split}
& L_g(x_v^{k},x_t^{k})=l^{x_v^{k}|x_t}+ l^{x_t^{k}|x_v}, \\ 
& l^{x_v^{k}|x_t}=-log\left ( \frac{exp(\mathcal{A}_g(x_v^{k},x_t^{k})/\tau)}{\sum_{j=1}^{N}exp(\mathcal{A}_g(x_v^{k},x_t^{j})/\tau)} \right ),\\ 
& l^{x_t^{k}|x_v}=-log\left ( \frac{exp(\mathcal{A}_g(x_v^{k},x_t^{k})/\tau)}{\sum_{j=1}^{N}exp(\mathcal{A}_g(x_v^{j},x_t^{k})/\tau)} \right ).
\end{split}
\label{eq:lext}
\end{equation}
Here, for a given $x_v^{k}$, $l^{x_v^{k}|x_t}$ aims to increase its similarity to its corresponding report and decrease its similarity to other reports ($j \neq k$). 
Similarly, for a given report $x_t^{k}$, $l^{x_t^{k}|x_v}$ aims to increase its similarity to its corresponding image and decrease its similarities  to other images in the batch.

\begin{figure}[tb]
\centering
\includegraphics[width=0.35\textwidth]{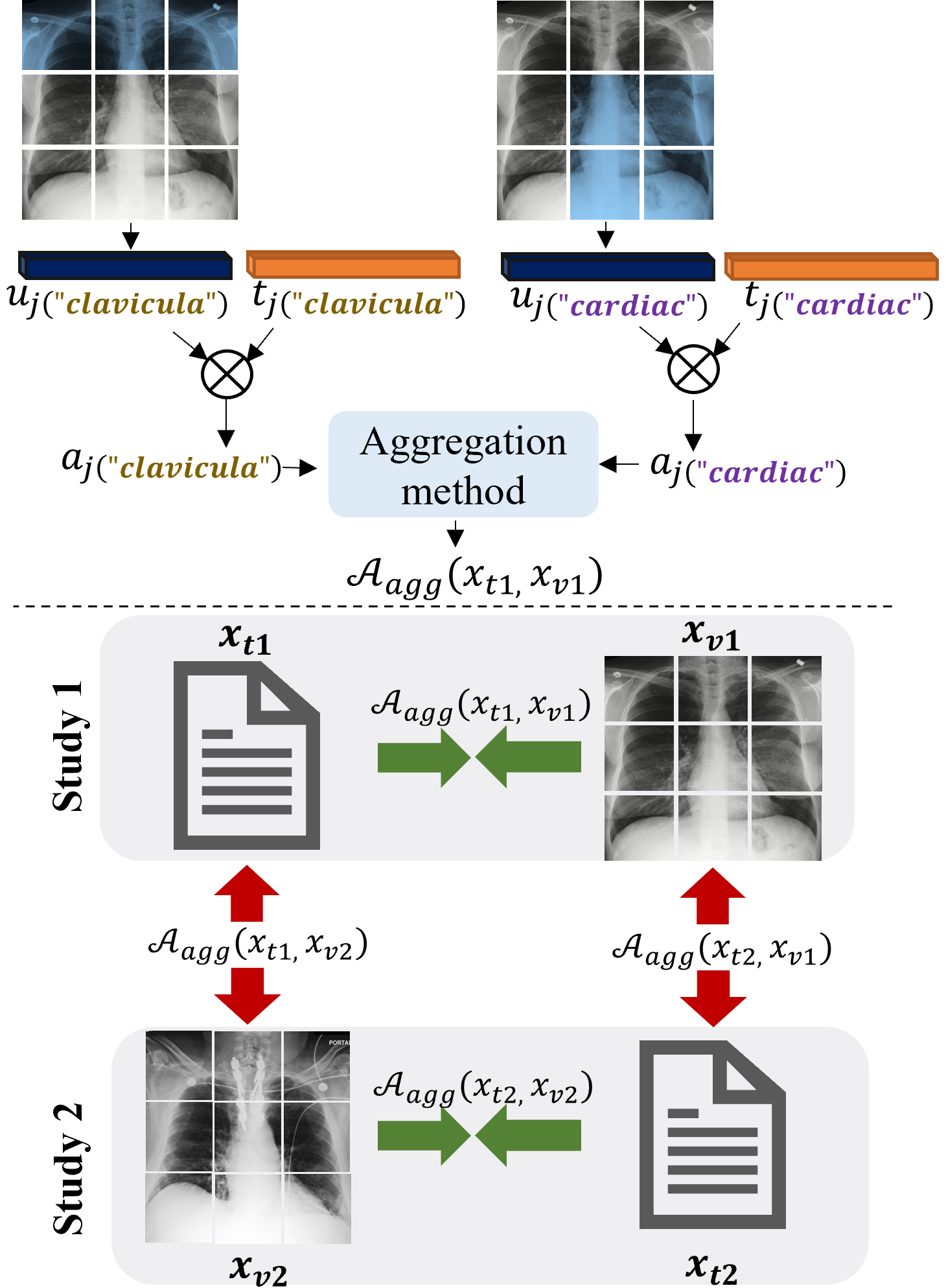}
\caption{{\bf Local external loss.} 
Top: Given a local textual representation $t_j$ (\textcolor{clavicula}{clavicula}) and its corresponding weighted visual representation $u_j$, we generate an aggregated alignment score $\mathcal{A}_{agg}$, based on the local alignments $\{ a_j \}$. 
Bottom:  $\mathcal{A}_{agg}$ is used to bring closer (\textcolor{arrow_green}{green}) the corresponding image-report pair $(x_{t1},x_{v1})$ and farther away (\textcolor{arrow_red}{red}) non-corresponding pairs e.g., $(x_{t1},x_{v2})$.}
\label{fig:local external loss}
\end{figure}

The local external loss, $L_{ext}$, aims to increase the similarity of positive pairs and decrease the similarity of negative ones, this time through the use of the local alignments. 
Thus, we use the aggregated local similarity score, $\mathcal{A}_{agg}$ (Section~\ref{subsec:aggregation}), instead of $\mathcal{A}_g$, as the objective for maximization and minimization, as illustrated in Figure~\ref{fig:local external loss}.

\begin{figure}[tb]
\centering
\includegraphics[width=0.74\linewidth]{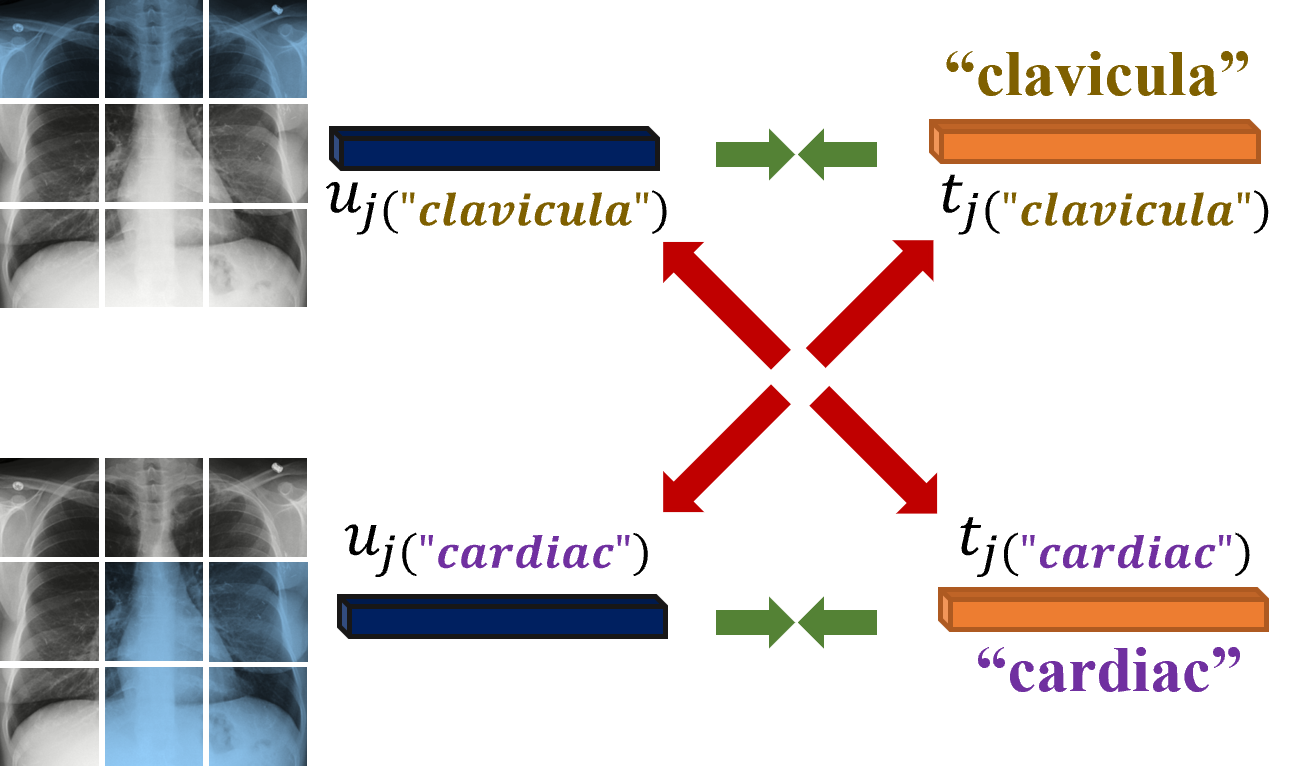}
\caption{{\bf Local internal loss.} 
Given local textual representations, $t_j$, and visual representations, $u_j$, this loss brings closer (\textcolor{arrow_green}{green}) corresponding representations, e.g.  $t_{j("clavicula")}$ and $u_{j("clavicula")}$ and further apart (\textcolor{arrow_red}{red}) non-corresponding representations, e.g. $t_{j("cardiac")}$ and $u_{j("clavicula")}$, from the same example.
}
\label{fig:local internal loss}
\end{figure}

The local internal loss, $L_{int}$, is given the local textual representation, $t_j$, as well as its corresponding attention weighted visual representation, $u_j$  from Equation~\ref{eq:uj}.
It aims to improve the local representations by maximizing the similarity between corresponding
pairs and minimizing the similarity between non-corresponding pairs of the same example.
The loss is defined as follows:
\begin{equation}
\begin{split}
& L_{int}(x_v^{k},x_t^{k})=\sum_{j=1}^{N_w}(l_{k}^{t_j|u}+ l_{k}^{u_j|t}), \\ 
& l^{t_j|u}_k=-log\left ( \frac{exp(a_j(t_j,u_j)/\tau)}{\sum_{i=1}^{N_w}exp(a_j(t_j,u_i)/\tau)} \right ),\\ 
& l^{u_j|t}_k=-log\left ( \frac{exp(a_j(t_j,u_j)/\tau)}{\sum_{i=1}^{N_w}exp(a_j(t_i,u_j)/\tau)} \right ).
\end{split}
\label{eq:lint}
\end{equation} 
Here, for a local textual representation $t_j$,  $l^{t_j|u}_k$ aims to increase the similarity between $t_j$ and its corresponding visual weighted representation, $u_j$, and to decrease the similarity to other
$u_{m \neq j}$ from the same study. 
The same rationale applies to $l^{u_j|t}_k$.
We calculate $l^{t_j|u}_k$ and $l^{u_j|t}_k$ separately, for each image-report pair $(x_v^{k},x_t^{k})$, as illustrated in Figure~\ref{fig:local internal loss}. 
Finally, we sum across all local pairs for a given study and then across all studies in the batch. 
Recall that the procedure described here is performed also for the visual representation in regards to the weighted textual representations.

\section{Experimental results}
\label{sec: experiments}

We examine our method on three retrieval applications:
text-image retrieval, phrase-grounding, and class-based retrieval.
For each application we compare against previous works that evaluate the specific application, as well as against additional methods that we trained.

\vspace{0.05in}
{\bf Datasets.}
We ran our experiments on three datasets: {\em MIMIC-CXR}, {\em CheXpert 5X200}, and {\em MS-CXR}.

(1)~{\em MIMIC Chest X-ray (MIMIC-CXR)} is a large, publicly available, dataset of chest radiographs, with free-text radiology reports~\cite{Johnson2019}. 
The dataset contains $377,110$ images, corresponding to $227,835$ radiographic studies performed at the Beth Israel Deaconess Medical Center. 
Each study in the dataset contains a report and one or more images of
{\em frontal} or {\em lateral}, where each study
 contains at least one frontal image. 
Each radiograph is associated with one or more classes, out of $14$ optional diagnostic labels. 
The two main sections of interest of the report are \textit{findings} and \textit{impression}.
Studies that do not contain these sections are filtered out. 
The data is randomly sampled to obtain validation and test sets, containing $1,000$ studies each.
The remaining training set contains $205,000$ studies, of which $100,000$ have both frontal and lateral images. 

(2)~{\em CheXpert 5X200}  
contains $200$ image-report pairs for $5$ abnormality categories~\cite{huang}, sampled from the CheXpert dataset~\cite{chexpert}. 
Each example in the dataset belongs to a single abnormality category.

(3)~{\em MS-CXR} is a subset of  MIMIC-CXR, which is extended for phrase-grounding~\cite{Boecking_2022}.
It contains labeled (text descriptions) bounding boxes for each image.
In total, there are $1,153$ pairs of region-text pairs.

\vspace{0.05in}
\noindent
{\bf Text-Image retrieval.}
Given a report, our goal is to find the most suitable image and vice versa.
This task evaluates how close matching pairs are in the feature space.
Hence, for this task, only the ground-truth image-report pairs are considered as positive. 
When both the frontal and the lateral images are available we use them both;  when only one exists we fill the missing image with zeros.
The accuracy of our model is measured by the $Recall@K$ metric, which returns the percentage of queries whose true match is successfully ranked within the top $K$ matches.

\begin{table}[tp]
\begin{tabular}{m{0.83in} | >{\centering}m{0.19in} >{\centering}m{0.16in} >{\centering}m{0.3in} | >{\centering}m{0.19in} >{\centering}m{0.16in} >{\centering\arraybackslash}m{0.3in}}
\multirow{2}{*}{Method} 
& \multicolumn{3}{c|}{\textbf{Image-to-Text}} & \multicolumn{3}{c}{\textbf{Text-to-Image}} \\
& R@1 & R@5 & R@10 & R@1 & R@5 & R@10 \\ \hline
MGCA~\cite{wang2022multi} & 25.8  & 51.9  & 62.1  & 27.9  & 51.2  & 61.6   \\
ConVIRT~\cite{Zhang2020} & 30.1  & 53.9   & 63.8   & 29.2  & 54.7 & 64.4 \\
GLoRIA~\cite{huang} & 30.3 & 57.5  & 66.5   & 24.0  & 51.8  & 62.8  \\ \hline
Ours \footnotesize{w/o LT\&PE} & \underline{36.1}  & \underline{59.1}  & \underline{69.1} & \underline{36.4} & \underline{60.7}  & \underline{70.5} \\
Ours  &  \textbf{39.7} &  \textbf{63.2}  & \textbf{71.7} & \textbf{37.7}&  \textbf{62.1} & \textbf{71.3}
\end{tabular}
\caption{\textbf{Text-Image retrieval results.}
Given a report, our goal is to retrieve the matching image and vice versa.
Our results outperform those of other methods on MIMIC-CXR, even without lateral (LT) images and structural information (PE).
Our full method further improves the results.
}
\label{table:retrieval}
\end{table}

\begin{table*}[t]
\centering
\begin{tabular}{m{0.82in} | >{\centering}m{0.5in} >{\centering}m{0.5in} >{\centering}m{0.5in} >{\centering}m{0.5in} >{\centering}m{0.5in} >{\centering}m{0.5in} >{\centering}m{0.5in} >{\centering}m{0.5in} >{\centering\arraybackslash}m{0.5in}}

Method & Atelectasis & Cardio- megaly & Consoli- dation & Lung Opacity & Edema & Pneumo- nia & neumo- thorax & Pleural Effusion & Average \\ \hline

ConVIRT~\cite{Zhang2020} & 0.86 & 0.64 & 1.25  & 0.78 & 0.68 & 1.03  & 0.28  & 1.02  & 0.818   \\
GLoRIA~\cite{huang}& 0.98  & 0.53  & 1.38 & 1.05 & 0.66 & 1.18 & 0.47 & 1.2 & 0.93     \\
BioViL~\cite{Boecking_2022}& \textbf{1.17} & 0.95 &\textbf{1.45} & 1.19 & 0.96  & 1.19 & 0.74 & \textbf{1.5} & 1.142    \\
Ours    &1.16&     \textbf{1.18} &   
1.37                  &    \textbf{1.37}  &         \textbf{1.05}         &   \textbf{1.27}   &    
\textbf{1.01}       &         1.24      &  
\textbf{1.206}
\end{tabular}
\caption{\textbf{Phrase-grounding results.}
Given a phrase and an image, the goal is to produce a similarity map between the phrase and the image. 
Our results outperform those of other methods, as reported in~\cite{Boecking_2022}, on MS-CXR. 
The results are measured using CNR; higher values indicate good localization of the phrase in the image. 
Qualitative results can be found in the supplementary materials.
}
\label{table: phrase grounding}
\end{table*}

Table~\ref{table:retrieval} compares our performance on MIMIC-CXR to
SoTA  methods for representation learning, ConVIRT~\cite{Zhang2020}, GLORIA~\cite{huang} and MGCA~\cite{wang2022multi}, which we have trained. 
Our method outperforms other methods in R@1, R@5 and R@10, both for the image-to-text task and for the text-to-image task, even without utilizing structural knowledge and lateral images.
Encoding domain structure and using the additional lateral images further improve the results.

\vspace{0.05in}
\noindent
{\bf Phrase-grounding.}
Given an image and a corresponding phrase, the goal is to produce an attention map of the similarity between the phrase and each image region~\cite{Boecking_2022}. 
The ground-truth is given as bounding boxes.
Hence, this task evaluates the local alignments. 
Following~\cite{Boecking_2022}, we measure the performance using {\em contrast to noise ratio (CNR)}, which measures the attention density inside the ground-truth bounding box.
High values indicate that the network accurately detects the regions of interest for the given phrase.
It is defined as: 
\begin{equation}
   CNR=\left | \mu _A-\mu_{\bar{A}} \right |/(\sigma_A^2+\sigma_{\bar{A}}^2)^{\frac{1}{2}},
\end{equation}
where $A$ and $\bar{A}$ are the interior and exterior of the bounding box, and $\mu$ and $\sigma$ are the mean and variance of the attention maps in each region. 

Table~\ref{table: phrase grounding} presents the results on the MS-CXR dataset.
The results are presented for each of the $8$ diagnostic categories, as well as for the average over the whole the dataset.
Our performance is compared to that of BioViL~\cite{Boecking_2022}, which introduced this task, and to the results of GLoRIA and ConVIRT that are reported in~\cite{Boecking_2022}.
Note that BioViL uses a different text encoder (CXR-bert) from all other methods in the table (Bioclinical-Bert). 
Our model achieves SoTA results, demonstrating the strength of our local representations and the localization ability of our model.

\vspace{0.05in}
\noindent
{\bf Class-based retrieval.}
Given an image, our goal is to retrieve reports that belong to the same class of the image.
For this task, a positive pair is defined as image-report that have the same abnormality label.
We follow~\cite{huang}'s settings, performing image-to-text retrieval and using the Precision@K metric to measure the accuracy of the retrieval.  
Precision@K is defined as the fraction of retrieved items (images) within the top $K$, which belong to the same class of abnormality as that of the query (textual descriptions).

Table~\ref{table:classbasedretrieval} shows that
our results outperform those of other methods in almost all metrics.
We note that~\cite{huang} was originally trained and evaluated on the  CheXpert dataset~\cite{chexpert}.
Since the reports of CheXpert are unavailable to the public, we trained on MIMIC-CXR~\cite{johnson2016mimic} and evaluated the retrieval in a zero-shot manner on the CheXpert 5X200 dataset, for all models.

\begin{table}[tb]
\centering
\begin{tabular}{l|lll}                                  \\
Method    & Prec@5         & Prec@10        & Prec@100         \\ \hline
ConVIRT~\cite{Zhang2020}   & \multicolumn{1}{c}{30.8} & \multicolumn{1}{c}{28.2} & \multicolumn{1}{c}{22.2} \\ 
GLoRIA~\cite{huang}   &     \multicolumn{1}{c}{32.6}                      &               \multicolumn{1}{c}{33.4}            &    \multicolumn{1}{c}{\textbf{29.0}}              \\
MGCA~\cite{wang2022multi}     &  \multicolumn{1}{c}{29.3}                       &    \multicolumn{1}{c}{27.6}                      &    \multicolumn{1}{c}{22.4}                      \\\hline
Ours     & \multicolumn{1}{c}{\textbf{37.2}}     & \multicolumn{1}{c}{\textbf{35.9}}     & \multicolumn{1}{c}{28.8}    
\end{tabular}
\caption
{\textbf{Class-based retrieval results.}
Given an image, reports that belong to the same class are retrieved. 
Our results outperform those of other methods on CheXpert 5X200 zero-shot evaluation. 
We note that this dataset contains only frontal views.}
\label{table:classbasedretrieval}
\end{table}

 \begin{figure*}[t]
 \centering
 \includegraphics[width=0.94\textwidth]{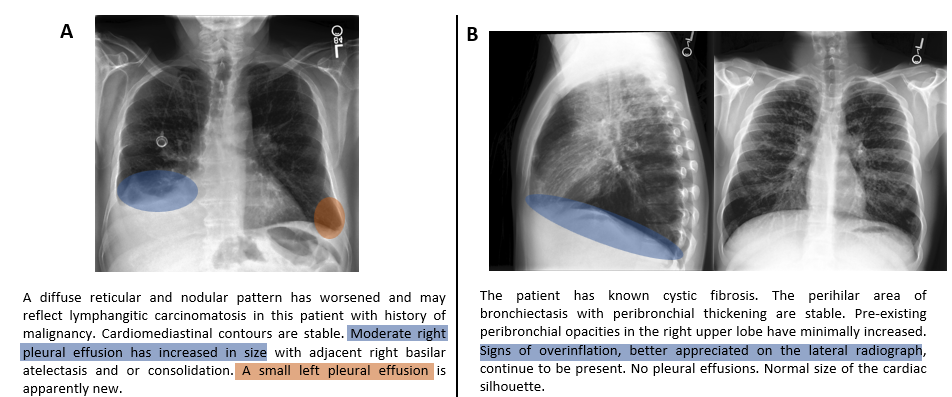}
\caption{{\bf Domain-knowledge importance.} 
(A) The positional information enables the model to consider the signs of right pleural effusion (\textcolor{teaser_blue}{blue}) and small left pleural effusion (\textcolor{teaser_brown}{orange}).
Thanks to our positional encoding, our model succeeds in the text-image retrieval task, but would fail otherwise.(B) The visual clue, the flattened diaphragm, which is a sign of over-inflation appears only in the lateral image (highlighted in \textcolor{teaser_blue}{blue}).
Thanks to  the lateral view, our model succeeds in the text-image retrieval task; it would fail if it used only the frontal image.}
\label{fig:ablation}
\end{figure*}

\section{Ablation study}
\label{sec:conclusion}
This section evaluates the contribution of the different components of our method. 

\vspace{0.05in}
\noindent
{\bf Losses.} 
Our model optimizes a combination of three losses: global, local-internal and local-external (Equation~\ref{eq:loss}). 
Table~\ref{table: ablation losses} demonstrates the contribution of each loss for both image-to-text and text-to-image retrieval tasks.
The best performance is achieved by the combination of the three losses. 
Thus, attempting to achieve the three goals of these losses indeed improves the learning process.

\begin{table}[tb]
\begin{tabular}{>{\centering}m{0.13in} >{\centering}m{0.13in} >{\centering}m{0.13in} | >{\centering}m{0.19in} >{\centering}m{0.16in} >{\centering}m{0.3in} | >{\centering}m{0.19in} >{\centering}m{0.16in} >{\centering\arraybackslash}m{0.3in}}
 &  &  & \multicolumn{3}{c|}{\textbf{Image-to-Text}} & \multicolumn{3}{c}{\textbf{Text-to-Image}}  \\ \hline
\textbf{$L_{int}$} & \textbf{$L_{ext}$} & \textbf{$L_{g}$} & R@1 & R@5 & R@10 & R@1 & R@5 & R@10 \\ \hline
                                      &                                     & $\checkmark$                                  &        31.8      &     58.2        &  66.8            &      33.6       &         58.0     &      68.0         \\
 $\checkmark$                                     &                                     &          $\checkmark$                         &      33.3    &    57.3      &    67.8       &   33.2      &       58.5   &   67.3        \\   
$\checkmark$                                     & $\checkmark$                                     &                                  &      33.2    &    58.4      &    68.0       &   31.5      &       57.9   &   67.9        \\
 $\checkmark$                                     &  $\checkmark$                                     &  $\checkmark$                                &    \textbf{36.1}          &      \textbf{59.1}      &      \textbf{69.1}       &    \textbf{36.4}        &       \textbf{60.7}       &      \textbf{70.5}       
\end{tabular}
\caption{\textbf{Loss ablation.} 
The best performance is achieved when combining all the three losses of Equation~\ref{eq:loss}. }
\label{table: ablation losses}
\end{table}

\noindent
{\bf Positional encoding.} 
Recall that we utilize domain-specific knowledge
by adding  positional encoding vectors.
Table~\ref{table: ablation domain knowledge} shows that positional encoding improves the results in most metrics.
It is beneficial both when using a single view or two views.
Figure~\ref{fig:ablation}(A) demonstrates the contribution of positional encoding qualitatively.
When the report contains relevant location information "moderate right pleural effusion" (blue) and "small left pleural effusion" (orange), without positional encoding, our base model might fail to align the report and the image.
However, when trained with positional encoding, the alignment is successful.

\begin{table}[tb]
\centering

\begin{tabular}{>{\centering}m{0.14in} >{\centering}m{0.14in} | >{\centering}m{0.22in} >{\centering}m{0.19in} >{\centering}m{0.33in} | >{\centering}m{0.22in} >{\centering}m{0.19in} >{\centering\arraybackslash}m{0.33in}}
 &  & \multicolumn{3}{c|}{\textbf{Image-to-Text}} & \multicolumn{3}{c}{\textbf{Text-to-Image}}  \\ \hline
LT & PE & R@1 & R@5 & R@10 & R@1 & R@5 & R@10 \\ \hline
 &  & 36.1         & 59.1         & 69.1          & 36.4         & 60.7         & 70.5         \\
 & $\checkmark$                                  &         37.7     &      61.5        &      70.7         &   34.7           &   60.9           &  70.5             \\
$\checkmark$ & & 38.1         & 61.6         & \textbf{72.1}         & 37.3         & 61.2         & \textbf{71.5}          \\
$\checkmark$ &  $\checkmark$ &      \textbf{39.7}        & \textbf{63.2}             &  71.7             &  \textbf{37.7}            & \textbf{62.1}             & 71.3             
\end{tabular}
\caption{{\bf Domain knowledge.} 
Adding either lateral (LT) images or visual structure information (PE) to the baseline frontal images improves the results in both tasks.
}
\label{table: ablation domain knowledge}
\end{table}

\vspace{0.05in}
\noindent
{\bf Lateral images.} 
Lateral images are explicitly mentioned in many reports, hence using them (when available) is desirable.
Table~\ref{table: ablation domain knowledge} confirms this, by showing superior results across all metrics, compared to using only the frontal images.
Figure~\ref{fig:ablation}(B) shows an example in which when our model is trained without lateral images, the alignment between the report and the image fails.
Using both views results in a correct alignment. 
The report mentions signs of overinflation that are "better appreciated on the lateral radiograph" (blue).
The signs for overinflation in the lateral image are more evident than in the frontal image.

\vspace{0.05in}
\noindent
{\bf Limitations.} 
Similarly to~\cite{Boecking_2022,lovt}, the limitation of our  approach is
that it does not explicitly deal with false negatives in the contrastive losses.
That is to say, there may be multiple reports that match a given image (and vice versa), but only one is considered positive.

\section{Conclusions}
This paper presented a new model for learning a joint X-ray image \& report representation. 
The model is based on a new alignment scheme that considers both local and global information. 
In addition, we propose to enrich the model with domain-specific information. 

The benefits of our representation is demonstrated on  
three types of retrieval tasks, two of which require large precision: text-image retrieval, phrase-grounding, and class-based retrieval.
Our model is shown to outperform SoTA models, even when the additional knowledge is unavailable.
The domain-specific knowledge adds to performance.

In the future, we would like to study loss functions that allow an image (/text) to be paired to multiple mates from the other domain.
For instance, the contrastive loss should not push away normal images from normal text of different pairs.
This has the potential to improve results across tasks and datasets, especially those of low diversity.

\newpage
{\small
\bibliographystyle{ieee_fullname}
\bibliography{egbib}

\begin{thebibliography}{10}\itemsep=-1pt

\bibitem{SIIM}
Society for imaging informatics in medicine: Siim-acr pneumothorax
  segmentation.

\bibitem{DBLP:journals/corr/abs-1904-03323}
Emily Alsentzer, John~R. Murphy, Willie Boag, Wei{-}Hung Weng, Di Jin, Tristan
  Naumann, and Matthew B.~A. McDermott.
\newblock Publicly available clinical {BERT} embeddings.
\newblock {\em CoRR}, abs/1904.03323, 2019.

\bibitem{Boecking_2022}
Benedikt Boecking, Naoto Usuyama, Shruthi Bannur, Daniel~C. Castro, Anton
  Schwaighofer, Stephanie Hyland, Maria Wetscherek, Tristan Naumann, Aditya
  Nori, Javier Alvarez-Valle, Hoifung Poon, and Ozan Oktay.
\newblock Making the~most of~text semantics to~improve biomedical
  vision{\textendash}language processing.
\newblock In {\em Lecture Notes in Computer Science}, pages 1--21. Springer
  Nature Switzerland, 2022.

\bibitem{sgraf}
Haiwen Diao, Ying Zhang, Lin Ma, and Huchuan Lu.
\newblock Similarity reasoning and filtration for image-text matching.
\newblock {\em CoRR}, abs/2101.01368, 2021.

\bibitem{vit}
Alexey Dosovitskiy, Lucas Beyer, Alexander Kolesnikov, Dirk Weissenborn,
  Xiaohua Zhai, Thomas Unterthiner, Mostafa Dehghani, Matthias Minderer, Georg
  Heigold, Sylvain Gelly, Jakob Uszkoreit, and Neil Houlsby.
\newblock An image is worth 16x16 words: Transformers for image recognition at
  scale.
\newblock {\em CoRR}, abs/2010.11929, 2020.

\bibitem{du2022survey}
Yifan Du, Zikang Liu, Junyi Li, and Wayne~Xin Zhao.
\newblock A survey of vision-language pre-trained models.
\newblock {\em arXiv preprint arXiv:2202.10936}, 2022.

\bibitem{resnet}
Kaiming He, Xiangyu Zhang, Shaoqing Ren, and Jian Sun.
\newblock Deep residual learning for image recognition.
\newblock {\em CoRR}, abs/1512.03385, 2015.

\bibitem{huang}
Shih-Cheng Huang, Liyue Shen, Matthew~P Lungren, and Serena Yeung.
\newblock {GLoRIA: A Multimodal Global-Local Representation Learning Framework
  for Label-efficient Medical Image Recognition}.

\bibitem{DBLP:journals/corr/abs-1712-02036}
Yan Huang, Qi Wu, and Liang Wang.
\newblock Learning semantic concepts and order for image and sentence matching.
\newblock {\em CoRR}, abs/1712.02036, 2017.

\bibitem{chexpert}
Jeremy Irvin, Pranav Rajpurkar, Michael Ko, Yifan Yu, Silviana Ciurea{-}Ilcus,
  Chris Chute, Henrik Marklund, Behzad Haghgoo, Robyn~L. Ball, Katie~S.
  Shpanskaya, Jayne Seekins, David~A. Mong, Safwan~S. Halabi, Jesse~K.
  Sandberg, Ricky Jones, David~B. Larson, Curtis~P. Langlotz, Bhavik~N. Patel,
  Matthew~P. Lungren, and Andrew~Y. Ng.
\newblock Chexpert: {A} large chest radiograph dataset with uncertainty labels
  and expert comparison.
\newblock {\em CoRR}, abs/1901.07031, 2019.

\bibitem{johnson2016mimic}
Alistair~EW Johnson, Tom~J Pollard, Lu Shen, Li-wei~H Lehman, Mengling Feng,
  Mohammad Ghassemi, Benjamin Moody, Peter Szolovits, Leo Anthony~Celi, and
  Roger~G Mark.
\newblock Mimic-iii, a freely accessible critical care database.
\newblock {\em Scientific data}, 3(1):1--9, 2016.

\bibitem{Johnson2019}
Alistair E.~W. Johnson, Tom~J. Pollard, Seth~J. Berkowitz, Nathaniel~R.
  Greenbaum, Matthew~P. Lungren, Chih-ying Deng, Roger~G. Mark, and Steven
  Horng.
\newblock {MIMIC-CXR, a de-identified publicly available database of chest
  radiographs with free-text reports}.
\newblock {\em Scientific Data}, 6(1):317, dec 2019.

\bibitem{DBLP:journals/corr/KarpathyF14}
Andrej Karpathy and Li Fei{-}Fei.
\newblock Deep visual-semantic alignments for generating image descriptions.
\newblock {\em CoRR}, abs/1412.2306, 2014.

\bibitem{DBLP:journals/corr/abs-1803-08024}
Kuang{-}Huei Lee, Xi Chen, Gang Hua, Houdong Hu, and Xiaodong He.
\newblock Stacked cross attention for image-text matching.
\newblock {\em CoRR}, abs/1803.08024, 2018.

\bibitem{DBLP:journals/corr/abs-1909-02701}
Kunpeng Li, Yulun Zhang, Kai Li, Yuanyuan Li, and Yun Fu.
\newblock Visual semantic reasoning for image-text matching.
\newblock {\em CoRR}, abs/1909.02701, 2019.

\bibitem{DBLP:journals/corr/abs-2004-06165}
Xiujun Li, Xi Yin, Chunyuan Li, Pengchuan Zhang, Xiaowei Hu, Lei Zhang, Lijuan
  Wang, Houdong Hu, Li Dong, Furu Wei, Yejin Choi, and Jianfeng Gao.
\newblock Oscar: Object-semantics aligned pre-training for vision-language
  tasks.
\newblock {\em CoRR}, abs/2004.06165, 2020.

\bibitem{Moon}
Jong~Hak Moon, Hyungyung Lee, Woncheol Shin, Young-Hak Kim, and Edward Choi.
\newblock {Multi-modal Understanding and Generation for Medical Images and Text
  via Vision-Language Pre-Training}.

\bibitem{lovt}
Philip M{\"{u}}ller, Georgios Kaissis, Congyu Zou, and Daniel Rueckert.
\newblock Joint learning of localized representations from medical images and
  reports.
\newblock {\em CoRR}, abs/2112.02889, 2021.

\bibitem{RSNA}
George Shih, Carol wu, Safwan Halabi, Marc Kohli, Luciano Prevedello, Tessa
  Cook, Arjun Sharma, Judith Amorosa, Veronica Arteaga, Maya
  Galperin-Aizenberg, Ritu Gill, Myrna Godoy, Stephen Hobbs, Jean Jeudy,
  Archana Laroia, Palmi Shah, Dharshan Vummidi, Kavitha Yaddanapudi, and Anouk
  Stein.
\newblock Augmenting the national institutes of health chest radiograph dataset
  with expert annotations of possible pneumonia.
\newblock {\em Radiology: Artificial Intelligence}, 1:e180041, 01 2019.

\bibitem{Smith-Bindman2008}
Rebecca Smith-Bindman, Diana~L. Miglioretti, and Eric~B. Larson.
\newblock {Rising Use Of Diagnostic Medical Imaging In A Large Integrated
  Health System}.
\newblock {\em Health Affairs}, 27(6):1491--1502, nov 2008.

\bibitem{DBLP:journals/corr/abs-1807-03748}
A{\"{a}}ron van~den Oord, Yazhe Li, and Oriol Vinyals.
\newblock Representation learning with contrastive predictive coding.
\newblock {\em CoRR}, abs/1807.03748, 2018.

\bibitem{vaswani2017attention}
Ashish Vaswani, Noam Shazeer, Niki Parmar, Jakob Uszkoreit, Llion Jones,
  Aidan~N Gomez, {\L}ukasz Kaiser, and Illia Polosukhin.
\newblock Attention is all you need.
\newblock {\em Advances in neural information processing systems}, 30, 2017.

\bibitem{wang2022multi}
Fuying Wang, Yuyin Zhou, Shujun Wang, Varut Vardhanabhuti, and Lequan Yu.
\newblock Multi-granularity cross-modal alignment for generalized medical
  visual representation learning.
\newblock {\em arXiv preprint arXiv:2210.06044}, 2022.

\bibitem{wang2020covid}
Linda Wang, Zhong~Qiu Lin, and Alexander Wong.
\newblock Covid-net: A tailored deep convolutional neural network design for
  detection of covid-19 cases from chest x-ray images.
\newblock {\em Scientific reports}, 10(1):1--12, 2020.

\bibitem{wang2022medclip}
Zifeng Wang, Zhenbang Wu, Dinesh Agarwal, and Jimeng Sun.
\newblock Medclip: Contrastive learning from unpaired medical images and text.
\newblock {\em arXiv preprint arXiv:2210.10163}, 2022.

\bibitem{objectCXR}
Zhiyun Xue, Sema Candemir, Sameer Antani, L. Long, Stefan Jaeger, Dina
  Demner-Fushman, and George Thoma.
\newblock Foreign object detection in chest x-rays.
\newblock pages 956--961, 11 2015.

\bibitem{Zhang2020}
Yuhao Zhang, Hang Jiang, Yasuhide Miura, Christopher~D. Manning, and Curtis~P.
  Langlotz.
\newblock {Contrastive Learning of Medical Visual Representations from Paired
  Images and Text}.
\newblock oct 2020.

\end{thebibliography}
}

\end{document}


\title{LIMITR: Leveraging Local Information for Medical Image-Text Representation (Supplementary material ) }
\author{First Author\\
Institution1\\
Institution1 address\\
{\tt\small firstauthor@i1.org}
\and
Second Author\\
Institution2\\
First line of institution2 address\\
{\tt\small secondauthor@i2.org}
}

\maketitle
\ificcvfinal\thispagestyle{empty}\fi
\section{Phrase-grounding}
For phrase-grounding evaluation (on MS-CXR dataset), we followed [3] and trained our model using only the impression section of the reports.
For this task, we dropped all the images of MS-CXR from the training set, to make sure that our model uses them only for inference.
Qualitative results presented below for each of the $8$ abnormality classes in MS-CXR dataset (Figure \ref{fig:phrasse grounding}).

\section{Cross modal alignment-extension }
Throughout the paper, for simplicity, we describe how our method creates and utilizes weighted visual representation with respect to each textual feature. We hereby extend the explanation from the paper and describe the complementary direction -- weighted textual features with respect to each image region. 

\textbf{Cross modal alignment.} 
This section is extension of section 3.2 from the paper. 
To create the weighted textual features with respect to each image region we start with computing the cosine similarity $c_{ij}$ between $v_{i}$ and $t_{j}$, to create $c_i=[c_{i1},c_{i2},\dots,  c_{iN_w}]$.
It is further
normalized using softmax,  in order to get an attention weight:
\begin{equation}
{w_{i}=softmax(\lambda c_{i})}.
\end{equation} 
The attended visual feature $m_i$ with respect to the $i^{th}$ image region is the weighted sum of all the textual local representations: 
\begin{equation}
{m_{i}=\sum_{j=1}^{N_w}w _{i}[j] \cdot t_{j}}.
\label{eq:uj}
\end{equation} 

Next, we calculate the alignment between $v_i$ and its corresponding  $m_i$. 
The local alignment $a_{i}$ is calculated as:
\begin{equation}
{a_{i}=\mathcal{A}(m_{i},v_{i})=\frac{ m_{i}\circ v_{i}}{\left \|  m_{i}\circ v_{i} \right \|_{2}}},
\label{eq:aj}
\end{equation}
where $\circ$ is an element-wise multiplication and $\left \| \cdot  \right \|_{2}$ is the $L_2$-norm.

\begin{figure}[tb]
\centering
\includegraphics[width=1\linewidth]{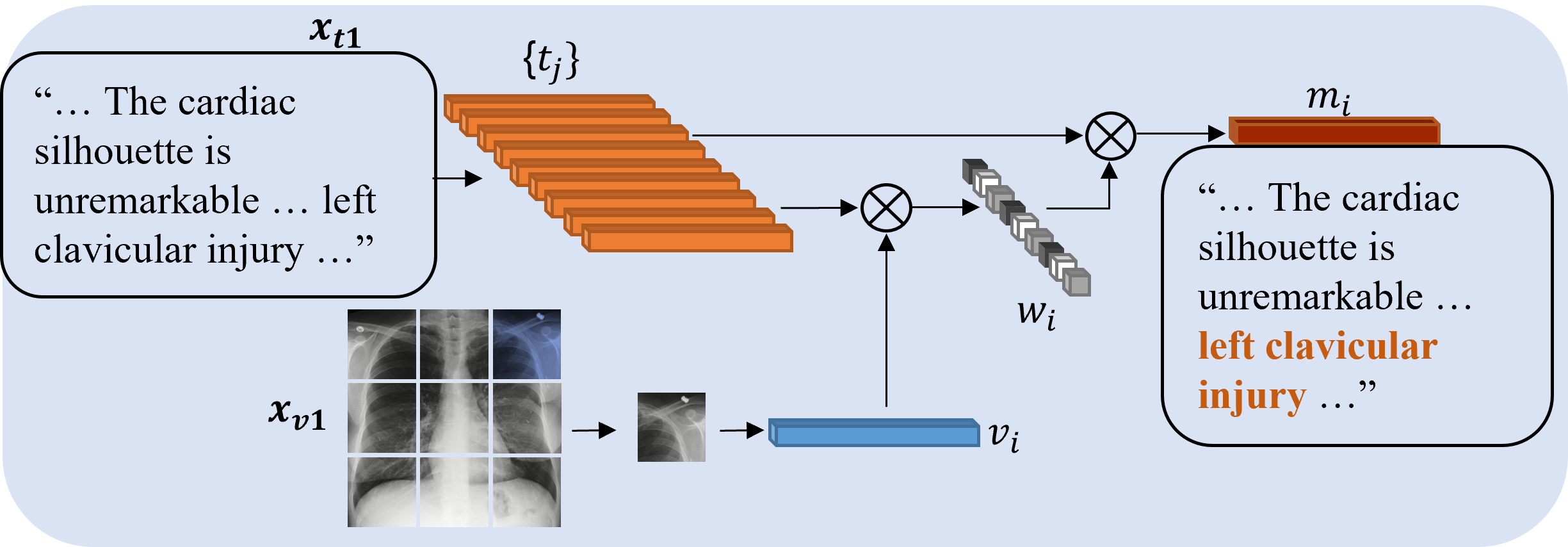}
\caption{{\bf Cross-modal alignment- textual weighted representation.} 
Given an image-report pair $(x_v,x_t)$, we compute for each image region $v_i$  its corresponding textual weighted representation $m_i$.
This is done by using the  similarity between each word representation $t_j$ to $v_i$ as the weight for this region $w_i[j]$. 
The right text shows that the words that correspond to the selected image region are highlighted in orange, representing the higher weight of these words.
The final textual representation, $m_i$, is created by a weighted sum of $t_j$. This figure corresponds to figure $3$ from the paper.
}
\label{fig:cross modal alignment}
\end{figure}

\textbf{Aggregation.} 
This section is extension of section 3.3 from the paper. 
The local alignments are aggregated into a single alignment vector using a weighted sum. 
Let the local alignments, computed at the alignment module, be $\mathcal{A}_{T}=\left \{a_{1},a_{2},a_{3}...a_{N_r}\right \}$.
Let $\bar{a}$ be the mean of $\mathcal{A}_T$.
The weight of $a_t$  is defined as:
\begin{equation}
q_{t}=\biggl( softmax \Bigl( \frac{W_q  \, \bar{a} \cdot (W_k \, \mathcal{A}_T)^T}{\sqrt{d}} \Bigr) \biggr)_t,
\end{equation} 
where $1 \leq t \leq N_r$, $W_q$ and $W_k$ are linear transformations of the self-attention, and $d$ is the feature dimension. 

The final alignment vector between an image and a report is defined as:
\begin{equation}
{a_{f}=\sum_{t=1}^{N_r} q _{t} \, (W_v \cdot a_{t})}.
\end{equation}
\textbf{Loss.}
This section is extension of section 3.4 from the paper. 
Recall that our loss is composed of three components: global, local internal and local external. 
The use of weighted textual features influence only local alignments, therefore we hereby describe the local internal and local external losses which use the weighted textual features with respect to each image region.

Local external loss: 
the loss is computed as described in equation (8), while this time $\mathcal{A}_{agg}$ is the aggregated representation based on the weighted-textual representations and the visual representations, as described in the section above.

Local internal loss:  $L_{int}$, is given the local visual representation, $v_i$, as well as its corresponding attention weighted textual representation, $m_i$.

The loss is defined as follows:
\begin{equation}
\begin{split}
& L_{int}(x_v^{k},x_t^{k})=\sum_{i=1}^{N_r}(l_{k}^{v_i|m}+ l_{k}^{m_i|v}), \\ 
& l^{v_i|m}_k=-log\left ( \frac{exp(a_i(v_i,m_i)/\tau)}{\sum_{j=1}^{N_r}exp(a_i(v_i,m_j)/\tau)} \right ),\\ 
& l^{m_i|v}_k=-log\left ( \frac{exp(a_i(v_i,m_i)/\tau)}{\sum_{j=1}^{N_r}exp(a_i(v_j,m_i)/\tau)} \right ).
\end{split}
\label{eq:lint}
\end{equation} 

We sum the losses related to the weighted visual features with the losses related to the weighted textual features for both the local internal loss and the local external loss.
\begin{figure*}[t]
\centering
\begin{tabular}{m{0.2in} P{1.5in} P{1.5in} P{1.5in} P{1.5in}}
\multirow{14}{*}{ \hfil\rotatebox[origin=c]{90}{\textbf{Atelectasis}}}&
 "Left lower lobe collapse is new" & "possible small left pleural effusion with adjacent atelectasis" & "left lower lobe is still collapsed" & "multisegmental lower lobe opacities are present, consistent with areas of atelectasis lung" \\
 &CNR 1.625 & CNR 1.932 & CNR 1.663 & CNR 0.909 \\
 &
\subfloat{\includegraphics[width=1\linewidth]{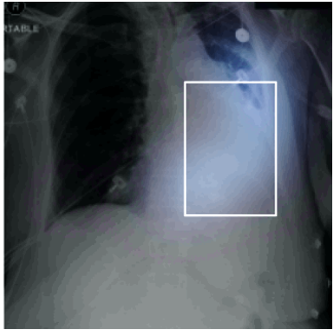}} &
\subfloat{\includegraphics[width=1\linewidth]{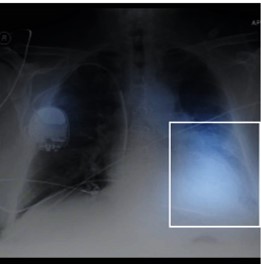}} &
\subfloat{\includegraphics[width=1\linewidth]{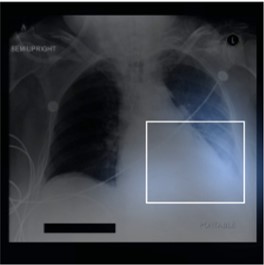}} &
\subfloat{\includegraphics[width=1\linewidth]{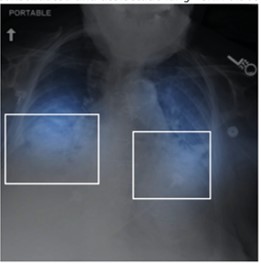}}\\\hline
\multirow{14}{*}{\hfil\rotatebox[origin=c]{90}{\textbf{Lung Opacity}}}&
 "hazy opacity in the left suprahilar lung may represent ground-glass opacity" & "patchy ground-glass opacities are seen in the left lung base" & "ground-glass opacities in the left lung" & "patchy ground-glass opacity in the mid right lung is also present" \\
 &CNR 1.877 & CNR 1.599 & CNR 1.413 & CNR 2.897 \\
 &
\subfloat{\includegraphics[width=1\linewidth]{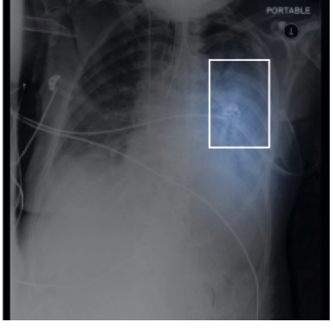}} &
\subfloat{\includegraphics[width=1\linewidth]{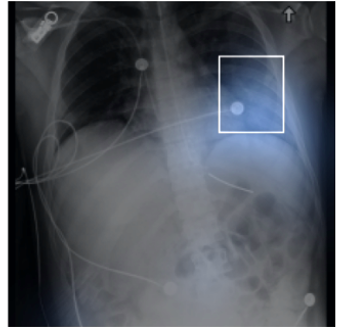}} &
\subfloat{\includegraphics[width=1\linewidth]{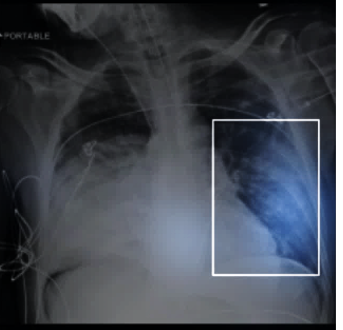}} &
\subfloat{\includegraphics[width=1\linewidth]{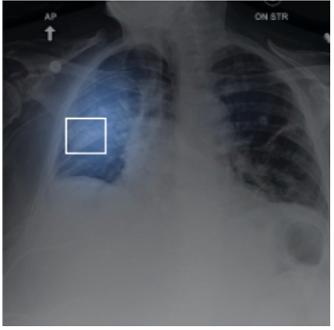}}\\\hline
\multirow{14}{*}{\hfil\rotatebox[origin=c]{90}{\textbf{Pleural effusion}}}&
 "small pleural effusion is stable" & "moderate right pleural effusion is unchanged compared to the prior exam" & "the minimal left pleural effusion persists" & "small amount of associated right pleural effusion is demonstrated" \\
&CNR 2.507 & CNR 1.900 & CNR 1.658 & CNR 1.602 \\
&
\subfloat{\includegraphics[width=1\linewidth]{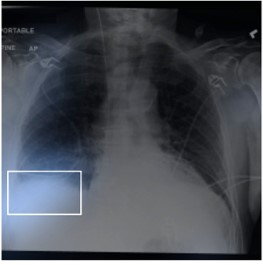}} &
\subfloat{\includegraphics[width=1\linewidth]{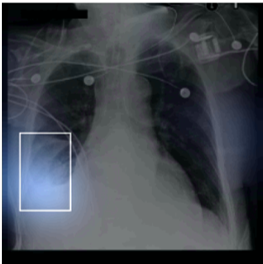}} &
\subfloat{\includegraphics[width=1\linewidth]{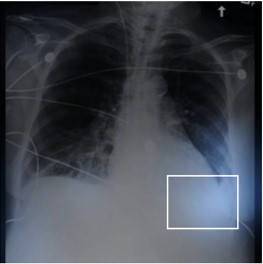}} &
\subfloat{\includegraphics[width=1\linewidth]{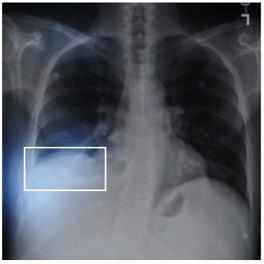}}\\\hline
\multirow{14}{*}{\hfil\rotatebox[origin=c]{90}{\textbf{Edema}}}&
 "evidence of worsening pulmonary edema and mitral regurgitation " & "hazy bilateral parenchymal opacities favored to represent edema" & "hazy perihilar opacities maybe due to pulmonary edema" & "interstitial edema is present in the right lower lung" \\
&CNR 1.832 & CNR 1.378 & CNR 1.698 & CNR 1.810 \\
&
\subfloat{\includegraphics[width=1\linewidth]{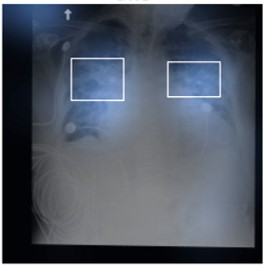}} &
\subfloat{\includegraphics[width=1\linewidth]{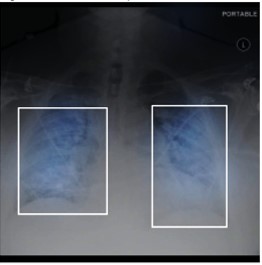}} &
\subfloat{\includegraphics[width=1\linewidth]{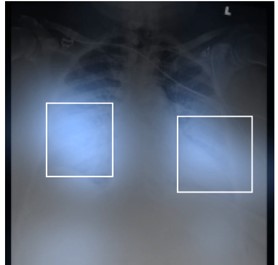}} &
\subfloat{\includegraphics[width=1\linewidth]{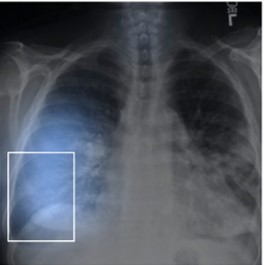}}\\\hline
\end{tabular}

\label{fig:phrasse grounding}
\end{figure*}

\setcounter{figure}{1}

\begin{figure*}[t]
\centering
\small
\begin{tabular}{m{0.2in} P{1.5in} P{1.5in} P{1.5in} P{1.5in}}
\multirow{14}{*}{ \hfil\rotatebox[origin=c]{90}{\textbf{Cardiomegaly}}}&
"enlarged cardiac silhouette" & "enlarged cardiac silhouette" & "enlarged cardiac silhouette" & "heart size is enlarged" \\
&CNR 1.455 & CNR 1.471 & CNR 1.294 & CNR 0.908 \\
&
\subfloat{\includegraphics[width=1\linewidth]{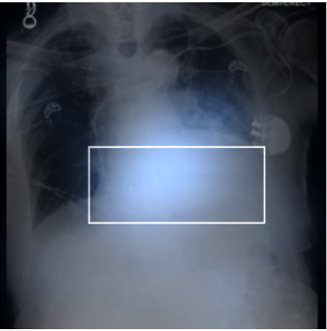}} &
\subfloat{\includegraphics[width=1\linewidth]{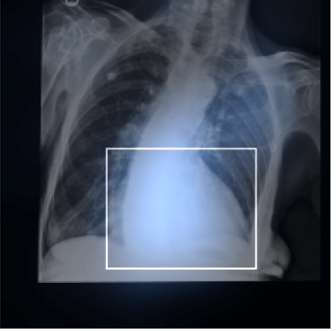}} &
\subfloat{\includegraphics[width=1\linewidth]{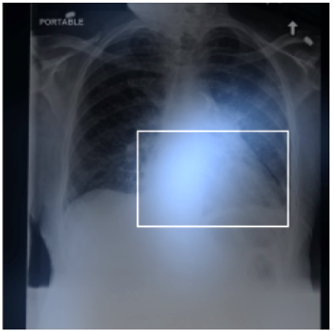}} &
\subfloat{\includegraphics[width=1\linewidth]{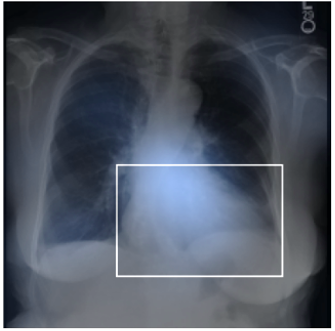}}\\\hline
\multirow{14}{*}{ \hfil\rotatebox[origin=c]{90}{\textbf{Pneumonia}}}&
"left lower lobe pneumonia" & "right lower lung pneumonia" & "patchy left base opacity" & "parenchymal opacity in the left lower lobe" \\
&CNR 2.538 & CNR 1.877 & CNR 2.097 & CNR 1.931 \\
&
\subfloat{\includegraphics[width=1\linewidth]{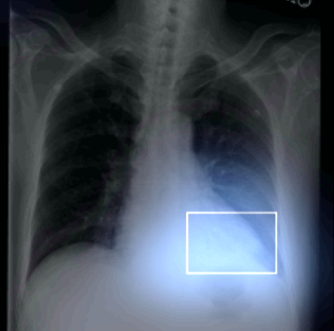}} &
\subfloat{\includegraphics[width=1\linewidth]{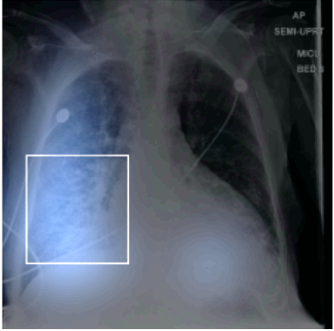}} &
\subfloat{\includegraphics[width=1\linewidth]{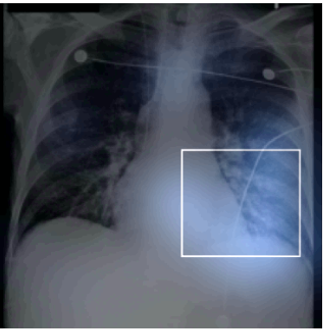}} &
\subfloat{\includegraphics[width=1\linewidth]{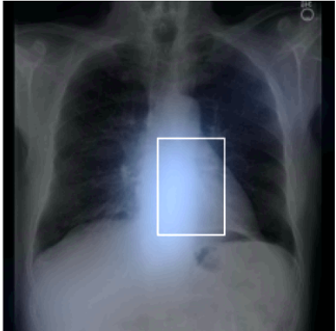}}\\\hline
\multirow{14}{*}{ \hfil\rotatebox[origin=c]{90}{\textbf{Pneumothorax}}}&
"large right pneumothorax" & "right-sided basal pneumothorax" & "right-sided hydro pneumothorax" & "apical component of pneumothorax" \\
&CNR 1.071 & CNR 1.385 & CNR 1.358 & CNR 1.370 \\
&
\subfloat{\includegraphics[width=1\linewidth]{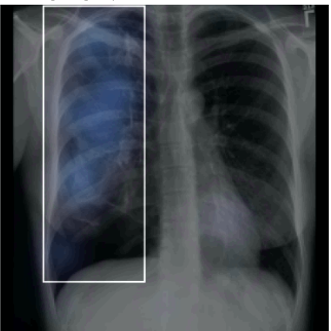}} &
\subfloat{\includegraphics[width=1\linewidth]{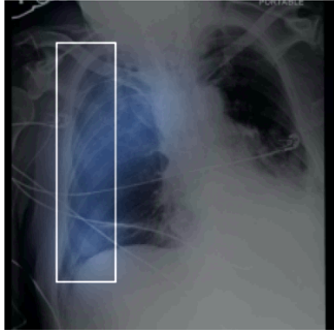}} &
\subfloat{\includegraphics[width=1\linewidth]{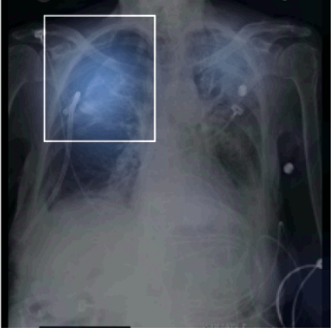}} &
\subfloat{\includegraphics[width=1\linewidth]{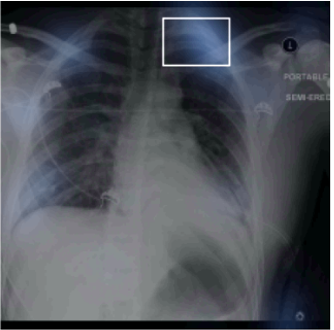}}\\\hline
\multirow{14}{*}{ \hfil\rotatebox[origin=c]{90}{\textbf{Consolidation}}}&
"there is left lower lobe consolidation" & "patchy consolidation in the mid left lung" & "patchy consolidation in the lower lung" & "patchy consolidation in the central right lung" \\
&CNR 1.982 & CNR 2.419 & CNR 2.118 & CNR 1.695 \\
&
\subfloat{\includegraphics[width=1\linewidth]{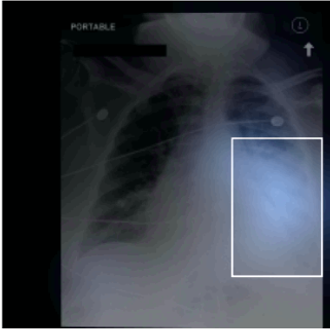}} &
\subfloat{\includegraphics[width=1\linewidth]{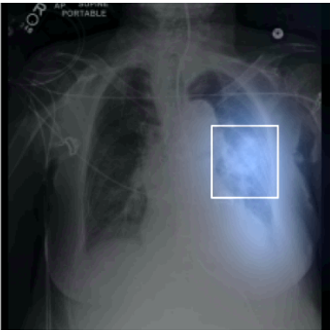}} &
\subfloat{\includegraphics[width=1\linewidth]{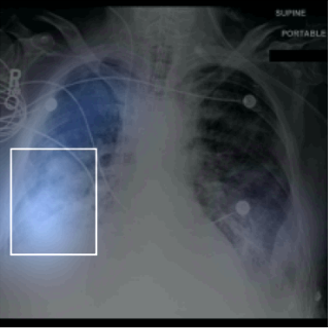}} &
\subfloat{\includegraphics[width=1\linewidth]{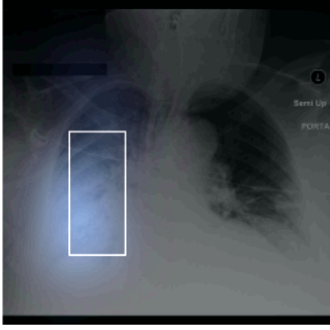}}\\\hline

\end{tabular}
\caption{\textbf{Phrase-grounding --- qualitative results.} Given a phrase and an image, the goal is to produce a similarity map between the phrase and the image. 
Brighter color indicates higher similarity of that region to the given phrase.
The results are measured using CNR; higher values indicate good localization of the phrase in the image.
Each row in the figure represents one of the abnormalitiy classes of MS-CXR dataset. 
Recall that our model was not trained for that task, but still learned to match well between image regions and their corresponding phrases.
}
\label{fig:phrasse grounding}
\end{figure*}

\section{Implementation details}
\textbf{Data pre-processing.} 
For training we used MIMIC-CXR dataset.

Images: we resize the original images to 256 pixels on the larger dimension and randomly/center cropped (training/inference) the resized images. 
We used Resnet50 as our image encoder. 
We extract the local features from the last convolution layer to get 19X19 feature maps which result in 361 local image regions.

Reports: we extract the impression and findings sections of the report using the official script provided in the github of MIMIC-CXR.
We used Bio-clinical BERT tokenizer to tokenize the extracted text.
Following [8] we set the maximum number of tokens per report to 97. 

\textbf{Training details.}
We used Adam optimizer with an initial learning rate of 5e-5 and a weight decay of 1e-6. 
We measured the performance on the validation set using $Recall@K$ where K=1,5,10 for both Image-to-Text retrieval and Text-to-Image. 
We set the maximum number of epochs to 50 and used the sum of those metrics $R_{sum}$ as the early stopping criteria (plateau over 5 epochs).
The best checkpoint is defined as the checkpoint with the highest $R_{sum}$.
 We use batch size of 48.
 All models were trained on a single A100 GPU.
For all losses we used $\tau = 0.1 $.